%% file: main.tex
\documentclass[10pt,twocolumn,letterpaper]{article}

\usepackage[pagenumbers]{cvpr} 

\input{preamble}
\definecolor{cvprblue}{rgb}{0.21,0.49,0.74}
\usepackage[pagebackref,breaklinks,colorlinks,allcolors=cvprblue]{hyperref}

\usepackage{amsmath,graphicx}
\usepackage{tabularx}
\usepackage{adjustbox}
\usepackage{url}
\usepackage{array}
\usepackage{multirow}
\usepackage{amssymb}
\usepackage{pifont}
\usepackage{pgfplots}
\usepackage{multirow}
\usepackage{booktabs}
\usepackage{makecell}
\usepackage{tikz}
\usepackage{tipa}
\usepackage{xcolor}
\usepackage{caption}
\usepackage{subcaption}
\usepackage{lipsum}
\usepackage[table]{xcolor}
\usepackage{arydshln}
\usepackage{algorithm}
\usepackage{algorithmic}

\usepackage{tcolorbox}
\usepackage{kotex}
\pgfplotsset{compat=1.18}
\definecolor{lightblue}{rgb}{0.93,0.95,1.0} 

\def\paperID{*****} 
\def\confName{CVPR}
\def\confYear{2026}

\title{MAD: Modality-Adaptive Decoding for Mitigating Cross-Modal  \\ Hallucinations in Multimodal Large Language Models}

\author{ Sangyun Chung, Se Yeon Kim, Youngchae Chee, and Yong Man Ro\thanks{Corresponding author.} \\
Integrated Vision Language Lab, KAIST, South Korea \\
{\tt\small \{jelarum, seyeon.kim, litcoderr, ymro\}@kaist.ac.kr}
}

\begin{document}
\maketitle
\input{sec/0_abstract}    
\input{sec/1_intro}
\input{figures/figure2}
\input{sec/2_related_work}

\input{sec/3_method}
\input{sec/4_experiment}
\input{sec/5_conclusion}
{
    \small
    \bibliographystyle{unsrtnat}
    \bibliography{main}
}

\input{sec/X_suppl}

\end{document}

%% file: sec/0_abstract.tex
\begin{abstract}
Multimodal Large Language Models (MLLMs) suffer from cross-modal hallucinations, where one modality inappropriately influences generation about another, leading to fabricated output. This exposes a more fundamental deficiency in modality-interaction control. To address this, we propose Modality-Adaptive Decoding (MAD), a training-free method that adaptively weights modality-specific decoding branches based on task requirements. 
MAD leverages the model's inherent ability to self-assess modality relevance by querying which modalities are needed for each task. The extracted modality probabilities are then used to adaptively weight contrastive decoding branches, enabling the model to focus on relevant information while suppressing cross-modal interference. 
Extensive experiments on CMM and AVHBench demonstrate that MAD significantly reduces cross-modal hallucinations across multiple audio-visual language models (7.8\% and 2.0\% improvements for VideoLLaMA2-AV, 8.7\% and 4.7\% improvements for Qwen2.5-Omni). Our approach demonstrates that explicit modality awareness through self-assessment is crucial for robust multimodal reasoning, offering a principled extension to existing contrastive decoding methods. Our code is available at \href{https://github.com/top-yun/MAD}{https://github.com/top-yun/MAD}
\end{abstract}
\vspace{-4mm}

%% file: sec/1_intro.tex
\section{Introduction}
\label{sec:intro}
Multimodal Large Language Models (MLLMs)~\cite{zhang2023, han2023onellm, yixuanundefined, han2023imagebind, zijiaundefined, sun2024video, Maaz2023VideoChatGPTTD, Lin2023VideoLLaVALU, wang2024internvideo2} have achieved remarkable progress in recent years. They move beyond text-only understanding to integrate diverse modalities such as vision, audio, and language. By jointly processing multiple sensory inputs, these models aim to emulate human-like multimodal perception. They enable a wide range of applications, from video question answering~\cite{yu2019activitynetqadatasetunderstandingcomplex, xiao2021nextqanextphasequestionansweringexplaining, wu2024longvideobenchbenchmarklongcontextinterleaved, zhou2025mlvubenchmarkingmultitasklong} to audio-visual scene understanding~\cite{geng2025longvalevisionaudiolanguageeventbenchmarktimeaware, yang2025audiocentricvideounderstandingbenchmark, liu2025valorvisionaudiolanguageomniperceptionpretraining}. This ability to reason across modalities marks a key step toward more holistic and comprehensive AI systems.

\input{figures/figure1}
Despite their impressive progress, MLLMs still face a critical challenge beyond single-modality hallucination problems~\cite{gunjal2023, li2023, zhou2023, guan2024hallusionbench, holyundefined}.
Conventional hallucinations typically involve generating incorrect or fabricated information within a single modality.
In contrast, multimodal settings introduce a more subtle yet harmful failure pattern: cross-modal hallucinations~\cite{sung2024avhbench, leng2024curse}, where one modality improperly influences content generation about another.
Figure~\ref{fig:1}, shows a clear example of visual hallucination and  video-driven audio hallucination occur simultaneously. The input video contains strong visual cues, such as a boat. When asked to describe both video and audio, the base model exhibits visual hallucination—stating that the person ``demonstrates how to use a fish finder and provides tips on catching fish.''
More critically, because the visual modality exerted an inappropriate influence it fabricates audio events that do not exist in the input, such as ``The sound of the man's voice and the splash of a fish jumping out of the water,'' even though such sounds do not exist in the input. Conversely, audio-driven visual hallucinations occur when auditory signals cause the model to invent corresponding visual events.

Unlike traditional hallucinations which distort content within a single modality, cross-modal hallucinations reveal a deeper failure of modality separation—where information inappropriately influences and corrupts the understanding of another. 
Consequently, addressing this challenge requires more than just mitigating single-modality hallucination. The model must also possess the modality appropriateness judgment capability—the ability to self-assess the reliability, information content, and contextual relevance of each input modality for a given task. This is crucial because the model must effectively determine which modality is important for a specific task and how to utilize the modalities together. This complexity, involving multiple intertwined requirements, is why cross-modal hallucination is inherently difficult to resolve. The problem stems from a fundamental failure in modality-interaction control, rather than merely inadequate representations within a single modality. Specifically, the failure is rooted in fundamental issues--assigning appropriate modality weights, suppressing misleading modalities, and preserving boundaries.

Recent attempts to mitigate hallucinations in vision-language models (VLMs)~\cite{Leng2023MitigatingOH, Kim2024CODECS, Wang2024MitigatingHI, huo2025selfintrospectivedecodingalleviatinghallucinations, chuang2023dola} using training-free decoding methods. Contrastive  Decoding~\cite{li2023contrastive} for vision-language models compares the output distribution of the original input with output distribution of a distorted input—typically created by adding noise or masking the visual modality. More recently, this paradigm has been extended to multi-modal settings: Audio-Visual Contrastive Decoding (AVCD)~\cite{jung2025avcd} applies similar contrastive mechanisms across both audio and visual modalities to reduce hallucinations in multi-modal language models.

However, these existing methods share a fundamental limitation: they are modality-agnostic, lacking awareness of task-specific modality requirements. 
Existing multimodal CD methods, designed for single-modality scenarios, assume hallucinations primarily stem from visual corruption.
While AVCD extends contrastive decoding to multiple modalities, it still applies uniform distortions across \textit{all} modalities without considering which modality is actually relevant to the given task.This uniform, non-adaptive approach is insufficient for tackling cross-modal interference. A static, modality-agnostic approach cannot dynamically block irrelevant inputs from causing cross-modal corruption.

In this paper, we address this limitation by introducing \textbf{Modality-Adaptive Decoding (MAD)}, a novel training-free method that overcomes modality-agnostic limitation. MAD is the first approach to explicitly determine the modality requirements of each task and dynamically adapt the contrastive decoding strategy to mitigate cross-modal hallucinations.

Our core innovation lies in self-assessing the importance of modality and employing task aware modality weights. We first prompt the multimodal model to self-assess the modality relevance on the given task, then extract modality weights that guide adaptive fusion of modality-specific contrastive distributions. This task-conditioned fusion enables the model to concentrate on relevant modality while suppressing inappropriate cross-modal interference.

We evaluate MAD on two comprehensive benchmarks for cross-modal hallucinations: The Curse of Multi-Modalities (CMM)~\cite{leng2024curse} and AVHBench~\cite{sung2024avhbench}. Experiments show that MAD effectively mitigates cross-modal hallucinations, achieving significant improvements on both benchmarks while maintaining performance on standard evaluation tasks.

Our main contributions are as follows:
\begin{itemize}[leftmargin=*,noitemsep,topsep=0pt]
    \item We propose a training-free, modality-adaptive decoding method that dynamically determines modality requirements for each task and adjusts contrastive decoding accordingly to mitigate cross-modal hallucinations in audio-visual LLMs.
    
    \item We introduce a task-driven modality weighting scheme that extracts explicit modality preferences and enables adaptive fusion of modality-specific contrastive distributions.
    
    \item We demonstrate significant improvements on cross-modal hallucination benchmarks (CMM and AVHBench), showing that explicit modality-aware fusion effectively reduces cross-modal hallucinations without requiring model retraining.
\end{itemize}



%% file: figures/figure1.tex
\begin{figure}[t]
\centering
\centerline{\includegraphics[width=1.0\linewidth]{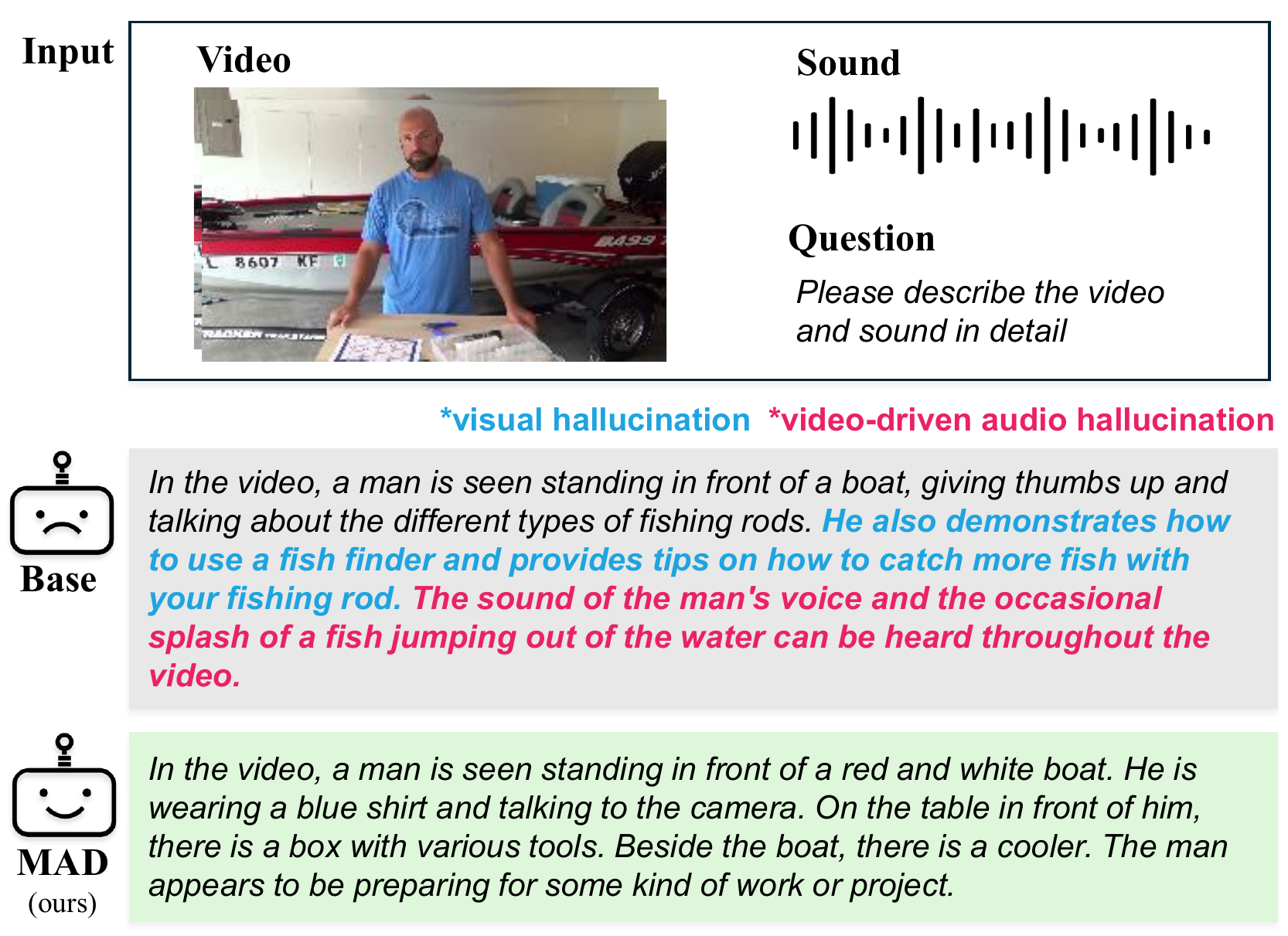}}
\vspace{-2mm}
\caption{\textbf{Cross-modal hallucinations and their mitigation through Modality-Adaptive Decoding (MAD).} The base model hallucinates non-existent visual content (red text) and audio events (blue text) when describing audio-visual inputs. MAD eliminates these hallucinations by adaptively suppressing cross-modal interference, producing accurate descriptions grounded in actual content.}
\label{fig:1}
\vspace{-4mm}
\end{figure}

%% file: figures/figure2.tex
\begin{figure*}[t]
\centering
\centerline{\includegraphics[width=1.0\linewidth]{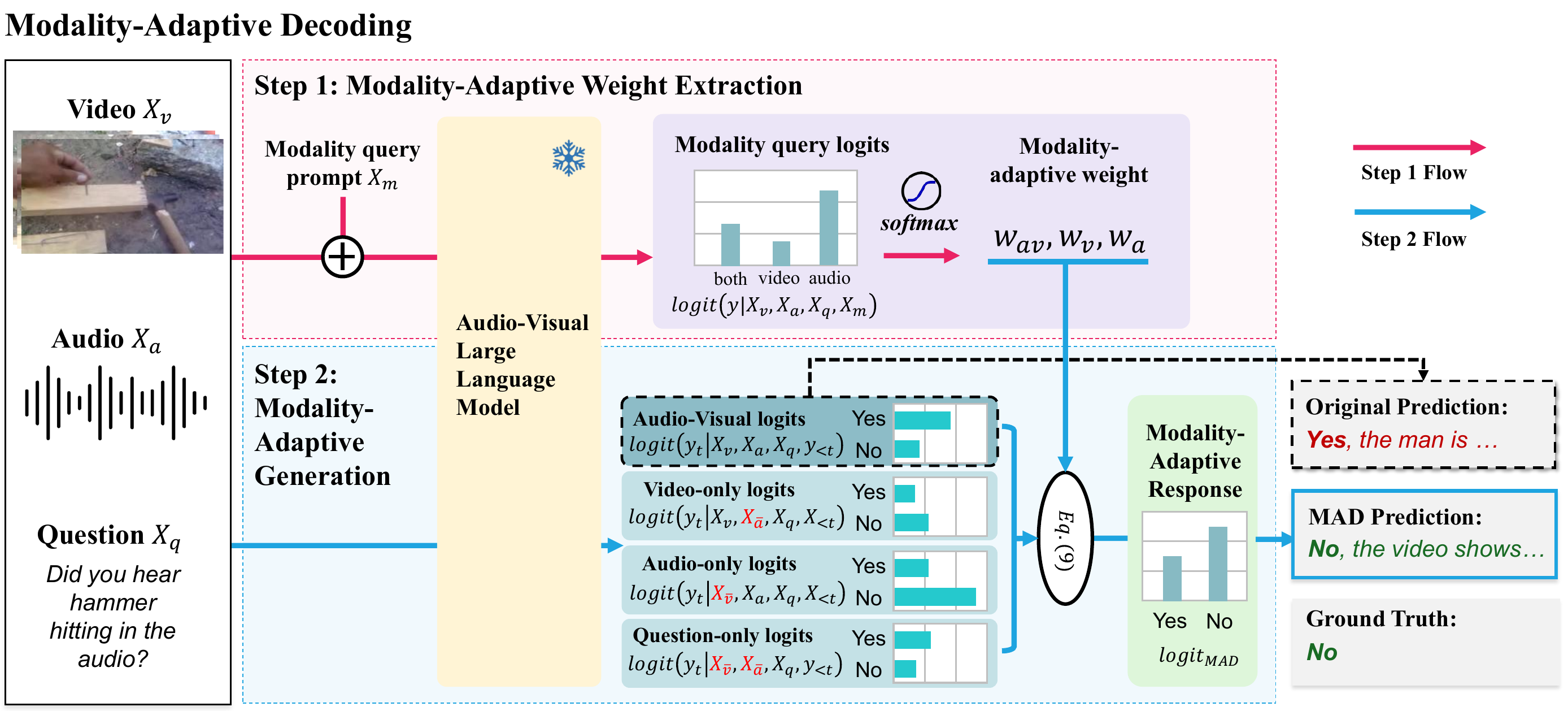}}
\vspace{-2mm}
\caption{\textbf{Overall MAD pipeline.} 
Given audio-visual inputs and a question, MAD extracts modality-adaptive weights by querying the model to identify relevant modalities [\textit{Step 1}]. 
During generation, MAD fuses contrastive logits computed from four modality configurations using these weights to dynamically emphasize relevant modalities [\textit{Step 2}].
In this example, despite a hammer being visible in the video, MAD correctly predicts ``No'' by prioritizing audio content, whereas the baseline predicts ``Yes'' due to cross-modal interference. This demonstrates MAD's effectiveness in mitigating video-driven audio hallucinations through adaptive, question-aware decoding.}
\label{fig:2}
\vspace{-4mm}
\end{figure*}

%% file: sec/2_related_work.tex
\section{Related Works}
\label{sec:related}
\subsection{Audio-visual Large Language Models.}
Recent progress in multimodal foundation models has improved beyond simply augmenting LLMs with visual perception to incorporating multiple modalities such as audio, vision and text~\cite{zhan2025anygptunifiedmultimodalllm, han2023onellm,ye2024catenhancingmultimodallarge, lyu2023macawllmmultimodallanguagemodeling, chen2022beatsaudiopretrainingacoustic, panagopoulou2024xinstructblipframeworkaligningxmodal, zhao2023chatbridgebridgingmodalitieslarge}. A growing line of work extends language models toward audio-visual understanding, enabling richer reasoning grounded in sound, image, and video. Video-LLaMa~\cite{zhang2023} enables end-to-end video understanding by jointly modeling visual and auditory signals within a LLM framework. Qwen2.5-omni~\cite{xu2025qwen2} perceives text, images, audio and video while generating real-time text and speech responses. These Audio-Visual LLMs (AV-LLMs) leverage multiple sensory streams to construct deeper representations of real-world events and improve semantic grounding.

\subsection{Mitigating hallucinations in MLLMs.}
Hallucination remains a major challenge for multimodal large language models (MLLMs), where generated responses diverge from input evidence. Existing approaches aim to reduce hallucination through (i) representation-level debiasing~\cite{jiang2024hallucination, zhang2024debiasing}, (ii) architectural decoding enhancements~\cite{xing2024mitigating, wang2024mllm, yang2025nullu, huang2024opera}, and (iii) training-free inference methods such as contrastive decoding~\cite{li2023contrastive, Leng2023MitigatingOH, Kim2024CODECS, Wang2024MitigatingHI, chuang2023dola, cho2025you}. Among these, training-free approaches are particularly compelling as they require no additional optimization or annotations.

Contrastive Decoding~\cite{li2023contrastive} generates outputs by contrasting likelihoods between a more plausible and a less plausible condition. Visual Contrastive Decoding (VCD)~\cite{Leng2023MitigatingOH} extends this idea by defining plausibility through visual distortions. CODE~\cite{Kim2024CODECS} leverages self-generated image descriptions as a contrastive reference to suppress hallucinatory tokens. ICD~\cite{Wang2024MitigatingHI} contrasts distributions obtained from standard versus disturbed instructions to remove hallucinated concepts during inference. DoLa~\cite{chuang2023dola} contrasts logits from later versus earlier language model layers, amplifying factual knowledge and reducing false predictions.

More recently, AVCD~\cite{jung2025avcd} extends contrastive decoding to audio-visual language models by identifying less-dominant modalities through attention and selectively perturbing them. AVCD~\cite{jung2025avcd} reformulates Contrastive Decoding~\cite{li2023contrastive} into a trimodal setting—audio, vision, and text—allowing hallucination suppression to operate on cross-modal interactions instead of a single corrupted modality. However, this method does not examine how varying levels of influence from each input modalities (e.g. audio and video) affect decoding. Our findings show that adaptively choosing the appropriate contrastive weight for each modality is crucial for further reducing hallucination.




%% file: sec/3_method.tex
\section{Method}
\label{sec:Method}
\subsection{Preliminaries}
\subsubsection{Multimodal Input and Generation Process}
An Audio-Visual Large Language Model (AV-LLM) takes three modalities as input: 
a video sequence $v$, an audio waveform $a$, and a textual question $q$.
Each modality is processed by its corresponding encoder specialized for that input type—
such as a visual encoder (e.g., CLIP~\cite{radford2021learningtransferablevisualmodels}, SigLIP~\cite{zhai2023sigmoidlosslanguageimage}) for video, an audio encoder (e.g., HuBERT~\cite{hsu2021hubertselfsupervisedspeechrepresentation}, Whisper~\cite{radford2022robustspeechrecognitionlargescale}) for audio, 
and a text tokenizer for language.
These modality-specific encoders transform raw inputs into tokenized representations 
that are compatible with a shared embedding space. We denote a sequence of input tokens of length $n$ as:
\[
    X = \{x_1, x_2, \ldots, x_n\},
\]
where each token $x_i$ represents an element in the multimodal input space. 
Specifically, the tokenized representations for each modality are given by:
\[
\begin{aligned}
    &X_v = \{x^v_1, \ldots, x^v_{n_v}\}, \\
    &X_a = \{x^a_1, \ldots, x^a_{n_a}\}, \\
    &X_q = \{x^q_1, \ldots, x^q_{n_q}\},
\end{aligned}
\]
where $n_v$, $n_a$, and $n_q$ denote the number of tokens for the video, audio, and text modalities, respectively.

The AV-LLM, parameterized by $\theta$, is denoted as $M_\theta$. 
At each decoding step $t$, the model autoregressively predicts the next-token distribution over the vocabulary $\mathcal{V}$.
The output vector $y_t \in \mathbb{R}^{|\mathcal{V}|}$ represents the unnormalized logits for each candidate token, 
from which the probability distribution is obtained through softmax normalization:
\begin{equation}
\begin{aligned}
y_t &\sim p_\theta(y_t \mid X_v, X_a, X_q, y_{<t}) \\
&\propto \text{logit}_\theta(y_t \mid X_v, X_a, X_q, y_{<t}),
\end{aligned}
\end{equation}
where $y_{<t} = \{y_1, \ldots, y_{t-1}\}$ denotes all previously generated tokens, 
and $\text{logit}_\theta(\cdot)$ represents the model’s raw output logits over $\mathcal{V}$ before softmax normalization.

\subsubsection{Contrastive Decoding for Vision Language Models}

Contrastive Decoding (CD) is a training-free inference method that mitigates hallucinations by contrasting the model’s output on a clean input with a output on a deliberately degraded input.
This naturally extends to vision–language models, where the degraded input is obtained by corrupting the image (e.g., noise, masking, or removal) to suppress visually induced hallucinations.

The key intuition is that visually grounded tokens exhibit a significant logit gap between the output from the clean input (with original vision $v$) and the output from the degraded input (with distorted or absent vision $\tilde{v}$). Conversely hallucinated tokens—arising from spurious correlations or language priors—maintain similar logits in both cases. By amplifying this contrast, the visual extension of CD suppresses hallucinations while preserving visually grounded generation.

Generally, given the vision and question input tokens $\{X_v, X_q\}$,  the visual extension of CD computes the contrastive logit for each candidate token $y_t$ as:
\begin{equation}
\begin{aligned}
\label{eq:cd_basic}
\text{logit}(y_t) &= (1+\alpha) \cdot \text{logit}(y_t \mid X_v, X_q, y_{<t}) \\
&\quad - \alpha \cdot \text{logit}(y_t \mid X_{\tilde{v}}, X_q, y_{<t})
\end{aligned}
\end{equation}
where $X_{\tilde{v}}$ denotes the vision input tokens with degraded vision $\tilde{v}$ obtained through techniques such as noise injection, augmentation or masking, and $\alpha > 0$ controls the contrastive strength. A larger $\alpha$ penalizes vision-independent tokens more heavily, while $\alpha = 0$ reduces to standard greedy decoding. By subtracting the scaled logits from the degraded branch, this process down-weights tokens relying on language priors, thereby mitigating visual hallucinations.

\subsection{Modality-Adaptive Decoding (MAD)}
In this section, we introduce Modality-Adaptive Decoding \textbf{(MAD)}. MAD extends and applies contrastive decoding to self-assess task-modality relevance and enable a task-aware distribution of modality weights.

\subsubsection{Weighted Contrastive Decoding}
Following the standard formulation of the visual extension of CD~\eqref{eq:cd_basic}, 
we extend the contrastive objective to an arbitrary modality $m$ as:
\begin{equation}
\begin{aligned}    
\label{eq:cd_modality}
\text{logit}_{\text{CD}}^{(m)}(y_t)
= \text{logit}(y_t \mid X_m, X_q, y_{<t}) + \alpha_m \cdot \Delta_m, \\
\Delta_m = \text{logit}(y_t \mid X_m, X_q, y_{<t})
- \text{logit}(y_t \mid X_{\tilde m}, X_q, y_{<t}).
\end{aligned}
\end{equation}
Here, $\Delta_m$ represents the contrastive signal measuring how much the model's output depends on modality $m$—larger values indicate stronger modality-specific contributions. The term $\alpha_m \cdot \Delta_m$ determines the strength of hallucination suppression: $\alpha_m$ controls how aggressively we penalize tokens that lack grounding in modality $m$.

However, conventional contrastive decoding applies a fixed $\alpha_m$ uniformly across all tasks, regardless of whether modality $m$ is actually relevant to the task. This is problematic: for a question like \textit{``What sound is heard?''}, we should apply strong contrastive suppression on the audio modality (high $\alpha_a$) while minimizing visual contrastive effects (low $\alpha_v$). Conversely, for \textit{``What color is the car?''}, the priorities reverse.

In other words, the contrastive strength $\alpha_m$ should be proportional to the importance of modality $m$ for the given task. To formalize this principle, we introduce a modality-adaptive weighting scheme:
\begin{equation}
\label{eq:adaptive_alpha}
\alpha_m = \gamma \cdot w_m
\end{equation}
where $\gamma > 0$ is a fixed base contrastive strength shared across all modalities, and $w_m \in [0,1]$ is a task-specific weight reflecting the relevance of modality $m$. By using a unified $\gamma$, we ensure that differences in contrastive intensity arise purely from the adaptive weights $w_m$, rather than from inherent modality-specific biases. This decomposition allows the contrastive intensity to scale dynamically: when $w_m$ is high (modality is relevant), $\alpha_m$ increases, amplifying hallucination suppression; when $w_m$ is low (modality is irrelevant), $\alpha_m$ decreases, reducing unnecessary penalization.

Substituting Eq.~\eqref{eq:adaptive_alpha} into Eq.~\eqref{eq:cd_modality}, we obtain the weighted contrastive decoding formulation:
\begin{equation}
\label{eq:macd}
\text{logit}_{\text{WCD}}^{(m)}(y_t)
= \text{logit}(y_t \mid X_m, X_q, y_{<t}) + \gamma \cdot w_m \cdot \Delta_m
\end{equation}

The key challenge now lies in determining $w_m$ dynamically for each question, which we address through a model self-assessment mechanism.



\subsubsection{Modality-Adaptive Weight Extraction}
A key idea of MAD is to let the multimodal model itself determine which modalities are most relevant to a given task.
To estimate the modality-specific weights $w_m$ in Eq.~\eqref{eq:macd}, we leverage the model’s inherent ability to self-assess modality relevance through the model prediction.
Given the inputs $X_v$, $X_a$, and $X_q$, we append a fixed modality query prompt $X_m$, instructing the model:
\textit{``To answer this question, which modality is needed (audio, video, or both)?''}

The model then autoregressively predicts the next token, from which we extract the logits corresponding to the tokens `video', `audio', and `both'.
These logits reflect the model’s confidence in each modality configuration:
\begin{equation}
    \begin{aligned}
        z_{av} &= \text{logit}(\text{`both'} \mid X_v, X_a, X_q, X_m), \\
        z_{v} &= \text{logit}(\text{`video'} \mid X_v, X_a, X_q, X_m), \\
        z_{a} &= \text{logit}(\text{`audio'} \mid X_v, X_a, X_q, X_m), \\
        [w_{av}, w_v, w_a] &= \text{softmax}([z_{av}, z_v, z_a]).
    \end{aligned}
\end{equation}
The resulting weights $(w_{av}, w_v, w_a)$ form normalized probabilities that quantify the model’s self-assessed importance of each modality.
These weights are subsequently used in Eq.~\eqref{eq:macd} to adaptively scale the contrastive decoding strength of each modality branch.

\input{figures/figure3}

\noindent\textbf{Analysis of Modality-Adaptive Weight.}
To verify whether the extracted modality weights $w_m$ accurately reflect the modality relevance implied by each task, we randomly sample 100 videos from VideoMME~\cite{fu2025videommefirstevercomprehensiveevaluation} and construct 300 questions categorized into three types: visual-related, audio-related, and audio-visual-related. For example, questions included \textit{'Is the folded paper white?'} (visual), \textit{'What kind of instrument is being played?'} (audio), and \textit{'Do hands move faster with the music?'} (audio-visual). More examples are in the supplementary material. For each question, we compute the averaged modality weights $(w_v, w_a, w_{av})$ obtained from Eq.~\eqref{eq:macd}.

As shown in Figure~\ref{fig:3}, the distribution of weights aligns well with intuitive modality dependencies:
visual-related questions yield dominant $w_v$, audio-related questions yield dominant $w_a$, and audio-visual questions exhibit dominant $w_{av}$ and a balanced combination of $w_v$ and $w_a$.
This observation confirms that the proposed modality query effectively captures modality relevance without any additional supervision, reinforcing the interpretability and reliability of our adaptive weighting scheme.


\subsubsection{Modality-Adaptive Generation}

Having established the weighted contrastive formulation in Eq.~\eqref{eq:macd} and the weight extraction mechanism, we now derive the complete generation process for audio-video multimodal inputs. 

To simplify notation, we define the following shorthand for logit terms based on modality configuration:
\begin{equation}
\label{eq:logit_notation}
\text{logit}^{m_1m_2\cdots} \triangleq \text{logit}(y_t \mid X_{m_1}, X_{m_2}, \ldots, X_q, y_{<t})
\end{equation}
where each modality in the superscript can be either standard ($m$) or perturbed ($\tilde{m}$). For brevity, we omit the token argument $y_t$ when clear from context.

Instantiating Eq.~\eqref{eq:cd_basic} for two modalities, video ($v$) and audio ($a$), and distinguishing between \emph{joint} audio–visual contrastive strength $\alpha_{av}$ and \emph{single-modality} strengths $\alpha_v$ and $\alpha_a$
leads to the following four-branch formulation:
\begin{equation}
\label{eq:avcd_decomposed}
\begin{aligned}
\text{logit}(y_t) 
&= \underbrace{(1 + \alpha_{av})\text{logit}^{vaq} - \alpha_{av} \cdot \text{logit}^{\tilde{v}aq}}_{\text{Visual CD $\mid$ audio present}} \\
&+ \underbrace{(1 + \alpha_{av})\text{logit}^{vaq} - \alpha_{av} \cdot \text{logit}^{v\tilde{a}q}}_{\text{Audio CD $\mid$ visual present}} \\
&+ \underbrace{(1 + \alpha_v)\text{logit}^{v\tilde{a}q} - \alpha_v \cdot \text{logit}^{\tilde{v}\tilde{a}q}}_{\text{Visual CD $\mid$ audio absent}} \\
&+ \underbrace{(1 + \alpha_a)\text{logit}^{\tilde{v}aq} - \alpha_a \cdot \text{logit}^{\tilde{v}\tilde{a}q}}_{\text{Audio CD $\mid$ visual absent}}.
\end{aligned}
\end{equation}
Each line implements a contrastive operation targeting a specific modality.
When both modalities are available (first two lines), we apply a joint
audio–visual contrastive decoding controlled by $\alpha_{av}$.
When one modality is absent (last two lines), the update falls back to
single-modality contrastive decoding with modality-specific strengths
$\alpha_v$ and $\alpha_a$.
In this way, the formulation aggregates four distinct contrastive signals,
each targeting hallucinations arising from a specific modality configuration.

Building on this four-branch formulation, we replace the fixed contrastive strengths
$\alpha_{av}$, $\alpha_v$, and $\alpha_a$ with the modality-adaptive terms
$\gamma \cdot w_{av}$, $\gamma \cdot w_v$, and $\gamma \cdot w_a$ derived in Eq.~\eqref{eq:adaptive_alpha}.
This substitution yields our proposed Modality-Adaptive Decoding (MAD):
\begin{equation}
\label{eq:mad_final}
\begin{aligned}
\text{logit}_{\text{MAD}}(y_t) 
&= \underbrace{(1 + \gamma \cdot w_{av})\text{logit}^{vaq} - \gamma \cdot w_{av} \cdot \text{logit}^{\tilde{v}aq}}_{\text{Visual CD $\mid$ audio present}} \\
&+ \underbrace{(1 + \gamma \cdot w_{av})\text{logit}^{vaq} - \gamma \cdot w_{av} \cdot \text{logit}^{v\tilde{a}q}}_{\text{Audio CD $\mid$ visual present}} \\
&+ \underbrace{(1 + \gamma \cdot w_v)\text{logit}^{v\tilde{a}q} - \gamma \cdot w_v \cdot \text{logit}^{\tilde{v}\tilde{a}q}}_{\text{Visual CD $\mid$ audio absent}} \\
&+ \underbrace{(1 + \gamma \cdot w_a)\text{logit}^{\tilde{v}aq} - \gamma \cdot w_a \cdot \text{logit}^{\tilde{v}\tilde{a}q}}_{\text{Audio CD $\mid$ visual absent}}.
\end{aligned}
\end{equation}

By adaptively scaling the contrastive strength for each modality according to question-specific importance, MAD more precisely suppresses hallucinations arising from irrelevant modalities while preserving information from relevant ones. The complete generation procedure is summarized in Algorithm~\ref{alg:mad}.

\input{tables/algorithm}



%% file: figures/figure3.tex
\begin{figure}[t]
\centering
\centerline{\includegraphics[width=1.0\linewidth]{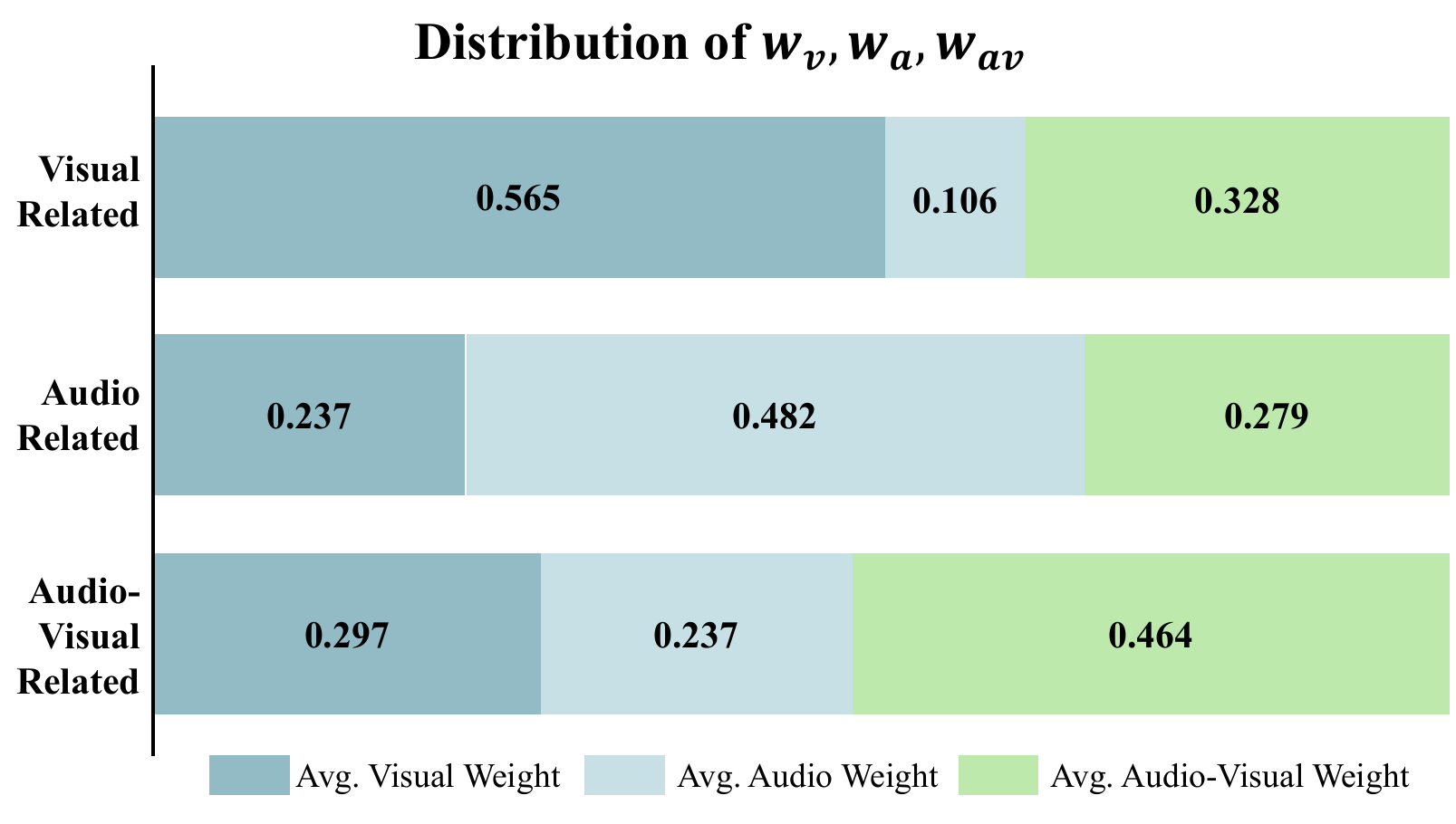}}
\vspace{-2mm}
\caption{Distribution of extracted modality weights $(w_v, w_a, w_{av})$ across question types on video. The weights align with intuitive modality dependencies, confirming that MAD correctly identifies the required modalities without supervision.}
\label{fig:3}
\vspace{-4mm}
\end{figure}

%% file: tables/algorithm.tex
\begin{algorithm}[H]
\caption{Modality-Adaptive Decoding (MAD)}
\label{alg:mad}
\begin{algorithmic}[1]
\REQUIRE Video input $X_v$, Audio input $X_a$, Question input $X_q$, Modality query prompt $X_m$, Audio-Visual LLM with parameters $\theta$, Contrastive strength $\gamma$
\ENSURE Decoded output sequence $y$

\STATE Initialize empty output sequence $y \leftarrow \emptyset$

\STATE Compute modality query logits:
\[
\text{logit}_m \leftarrow p_\theta(y|X_v, X_a, X_q, X_m)
\]
\STATE Extract modality-adaptive weights:
\begin{align*}
z_{av} &\leftarrow \text{logit}_m[\text{`both'}],  \\
z_v &\leftarrow \text{logit}_m[\text{`video'}], \\
z_a &\leftarrow \text{logit}_m[\text{`audio'}] \\
w_{av}, w_v, w_a &\leftarrow \text{softmax}([z_{av}, z_v, z_a])
\end{align*}

\WHILE{EOS token $\notin y$}
    \STATE Compute logits for all modality configurations:
    \begin{align*}
        \text{logit}^{vaq} &\leftarrow p_\theta(y_t \mid X_v, X_a, X_q, y_{<t}) \\
        \text{logit}^{\tilde{v}aq} &\leftarrow p_\theta(y_t \mid X_{\tilde{v}}, X_a, X_q, y_{<t}) \\
        \text{logit}^{v\tilde{a}q} &\leftarrow p_\theta(y_t \mid X_v, X_{\tilde{a}}, X_q, y_{<t}) \\
        \text{logit}^{\tilde{v}\tilde{a}q} &\leftarrow p_\theta(y_t \mid X_{\tilde{v}}, X_{\tilde{a}}, X_q, y_{<t})
    \end{align*}
    
    \STATE Compute MAD logits using Eq.~\eqref{eq:mad_final}:
    \small{
    \begin{align*}
        \text{logit}_{\text{MAD}}(y_t) &= (1 + \gamma \cdot w_{av})\text{logit}^{vaq} - \gamma \cdot w_{av} \cdot \text{logit}^{\tilde{v}aq} \\
        &+ (1 + \gamma \cdot w_{av})\text{logit}^{vaq} - \gamma \cdot w_{av} \cdot \text{logit}^{v\tilde{a}q} \\
        &+ (1 + \gamma \cdot w_v)\text{logit}^{v\tilde{a}q} - \gamma \cdot w_v \cdot \text{logit}^{\tilde{v}\tilde{a}q} \\
        &+ (1 + \gamma \cdot w_a)\text{logit}^{\tilde{v}aq} - \gamma \cdot w_a \cdot \text{logit}^{\tilde{v}\tilde{a}q}
    \end{align*}
    }
    \STATE Select next token: $\hat{y}_t \leftarrow \arg\max_{y_t \in \mathcal{V}} \text{logit}_{\text{MAD}}(y_t)$
    \STATE Append $\hat{y}_t$ to $y$
\ENDWHILE

\RETURN $y$
\end{algorithmic}
\end{algorithm}

%% file: sec/4_experiment.tex
\section{Experiments}
\input{tables/table1}
We evaluate the effectiveness of MAD on cross-modal hallucination benchmarks using multiple state-of-the-art audio-visual language models, comparing against existing contrastive decoding methods.
\subsection{Experimental Setup}

\subsubsection{Benchmark Datasets.}
We evaluate our method on two benchmarks designed to assess cross-modal hallucinations in multimodal large language models.
\textbf{AVHBench}~\cite{sung2024avhbench} is specifically designed to evaluate audio-visual hallucinations, containing video clips paired with questions that require reasoning about either visual or auditory information.
\textbf{CMM}~\cite{leng2024curse} provides a comprehensive evaluation of multimodal understanding, particularly examining how models depend on or bias toward certain modalities under diverse audio-visual scenarios.
Both benchmarks enable systematic assessment of how models handle modality-specific queries and resist cross-modal interference.

\subsubsection{Models.}
We conduct experiments on three state-of-the-art audio-visual multimodal large language models: VideoLLaMA2-AV-7B~\cite{cheng2024videollama} and Qwen2.5-Omni-7B~\cite{xu2025qwen2}.
These models represent diverse architectural approaches to multimodal understanding and provide a comprehensive testbed for evaluating cross-modal hallucination mitigation methods.

\subsubsection{Baselines.}
We compare our Modality-Adaptive Decoding (MAD) against existing contrastive decoding methods.
\textbf{VCD-Extended} applies the Visual Contrastive Decoding~\cite{Leng2023MitigatingOH} principle to the multimodal setting by contrasting against all possible modality distortions:
\begin{equation}
\begin{aligned}
\label{vcd_extended}
\operatorname{logit_{VCD-extended}}(y_t) &= (1 + 3\alpha) \cdot \operatorname{logit}^{vaq} - \alpha  \cdot \operatorname{logit}^{\tilde{v}aq} \\
&\quad - \alpha  \cdot \operatorname{logit}^{v\tilde{a}q}  - \alpha \cdot  \operatorname{logit}^{\tilde{v}\tilde{a}q}.
\end{aligned}
\end{equation}
\textbf{AVCD}~\cite{jung2025avcd} extends contrastive decoding to audio-visual settings but applies uniform distortion without query-specific adaptation.
These baselines represent the current state-of-the-art in training-free hallucination mitigation for multimodal models.

\subsubsection{Implementation Details.}
For all experiments, we set the temperature to 0 for deterministic generation across all models.
We determined the optimal values by sampling 100 examples per dataset and varying the strength hyperparameter $\gamma$ from 0.5 to 3.0 in 0.5 intervals.
Consequently, we set the strength hyperparameter $\gamma$ to 2.5 for all datasets.
All experiments are conducted using the same computational environment and random seeds to ensure reproducibility.

\subsection{Main Results}

Table~\ref{tab:main_results} presents the main experimental results comparing MAD with baseline methods across the CMM and AVHBench benchmarks.
Across all models and datasets, MAD consistently outperforms existing decoding methods, demonstrating its effectiveness in mitigating cross-modal hallucinations.

\noindent\textbf{Results on CMM.}
CMM evaluates hallucinations stemming from overreliance on unimodal priors, and MAD achieves clear gains across all three dominance categories.
Visual dominance\textbf{ (Visual Dom.)} occurs when a model over-relies on visual information while failing to properly incorporate linguistic and auditory cues.
Audio dominance\textbf{ (Audio Dom.)} occurs when a model places excessive emphasis on auditory input while failing to properly integrate visual or linguistic information.
Language dominance\textbf{ (Language Dom.)} appears when models follow linguistic priors even when they conflict with multimodal evidence.

For VideoLLaMA2-AV~\cite{cheng2024videollama}, MAD improves visual dominance by +9.3\% and language dominance by +5.5\%, resulting in an overall accuracy of 81.3\%.
For Qwen2.5-Omni~\cite{xu2025qwen2}, the improvements are even larger: visual dominance increases by +12.3\% and audio dominance by +12.0\%, bringing the overall accuracy to 81.4\%.
These consistent gains show that MAD effectively mitigates unimodal-prior-induced hallucinations by accurately identifying which modality each task requires.

\noindent\textbf{Results on AVHBench.}
MAD also delivers substantial improvements on AVHBench.
For VideoLLaMA2-AV, video-driven audio hallucination accuracy increases by +4.0\%, while Qwen2.5-Omni achieves a +5.7\% gain.
Audio-driven video hallucination further improves by +3.7\% for Qwen2.5-Omni, confirming that MAD reduces both audio-driven video and video-driven audio hallucinations.

Compared to VCD-Extended and AVCD, which apply uniform distortion strategies without query-specific adaptation, MAD’s adaptive weighting of modality-specific contrastive branches enables more precise suppression of dominance-induced hallucinations.
The consistent performance improvements across models and benchmarks underscore the importance of explicit modality-aware decoding for addressing cross-modal hallucinations in multimodal large language models.

\input{tables/table2_1}

\subsection{Ablation Studies}

We conduct ablation studies to analyze the contribution of individual components in MAD.

\subsubsection{Comparison of Weighting Strategies.}
To examine how different weighting strategies interact with the branch structure of Eq.~\eqref{eq:mad_final}, we compare MAD with two simplified variants: (1) a \emph{uniform weighting} scheme that assigns equal weights to all modalities ($w_{av}=w_v=w_a=\frac{1}{3}$), and (2) an \emph{argmax weighting} scheme that computes only the branch corresponding to the most relevant modality, identified by $\text{argmax}_m w_m$.
In Eq.~\eqref{eq:mad_final}, MAD is decomposed into four contrastive branches depending on the presence or absence of each modality.
The argmax variant activates only one of these branches (e.g., \textit{Video CD—audio absent} when the video modality is dominant), effectively skipping the others.
As shown in Table~\ref{tab:ablation}, both the uniform and argmax variants lead to clear performance degradation.
The uniform weighting overlooks query-specific modality demands, while the argmax strategy eliminates complementary cues from non-selected modalities.
In contrast, our adaptive weighting softly integrates all branches in Eq.~\eqref{eq:mad_final} according to their estimated relevance, achieving a balanced fusion that consistently outperforms both baselines across models and benchmarks.

\subsubsection{Contribution of Modality-Specific Weights.}
\input{tables/table4}

Table~\ref{tab:ablation_weight} reports the effect of removing each modality-specific weight in MAD.
We selectively ablate $w_a$, $w_v$, and $w_{av}$ to assess their individual contributions to mitigating cross-modal hallucinations.

Removing the audio weight ($w_a$) leads to a notable drop to 78.0\% accuracy, with a 6.5\% decrease in Visual Dominance.
Without $w_a$, the model fails to capture essential audio cues, which in turn increases its reliance on visual information and ultimately causes it to hallucinate audio events purely from visual signals (similar to the example in Figure~\ref{fig:2}).

Eliminating the visual weight ($w_v$) yields 78.3\% accuracy and a 3.0\% decline in Audio Dominance.
Here, the model overweights audio input and insufficiently incorporates visual evidence, limiting its ability to suppress audio-driven interference.
The symmetric degradation observed in both cases confirms that $w_a$ and $w_v$ are jointly essential for preventing dominance-induced hallucinations.

Although using both $w_a + w_v$ reduces the performance drop compared to single-weight ablations, noticeable drops remain in all three dominance. This configuration suppresses visual dominance on audio-related queries and audio dominance on visual-related ones, yet it still struggles on tasks requiring joint audio-visual reasoning.
To address this, $w_{av}$ provides a flexible coupling mechanism between the two modalities, enabling effective fusion when joint reasoning is necessary. When all three weights ($w_a + w_v + w_{av}$) are used, MAD achieves the highest accuracy of 81.3

Overall, these results highlight a key insight: cross-modal hallucinations arise bidirectionally-visual dominance induces audio hallucinations, and audio dominance induces visual ones.
Effective mitigation therefore requires adaptively balancing modalities based on each query’s modality relevance.

\subsubsection{Performance on General Audio-Visual QA Tasks.}
\input{tables/table3}
To evaluate whether MAD also benefits standard audio–visual question answering tasks, we additionally test on general AVQA benchmarks that do not explicitly target cross-modal hallucinations.
As shown in Table~\ref{tab:general_QA}, MAD achieves comparable or slightly improved performance across these tasks.
We attribute this improvement to MAD’s ability to suppress subtle hallucination behaviors that can mislead the model even in non-hallucination-focused benchmarks.
These results suggest that adaptive modality selection not only mitigates explicit cross-modal hallucinations but also enhances general AVQA performance by guiding the model toward more reliable evidence.

%% file: tables/table1.tex
\begin{table*}[t]
\centering
\renewcommand{\arraystretch}{1.0}
\renewcommand{\tabcolsep}{2.0mm}
\resizebox{1.0\linewidth}{!}{
\begin{tabular}{lccccccc}
\toprule
\multirow{3.5}{*}{\textbf{Model}} &
\multicolumn{4}{c}{\textbf{CMM}} &
\multicolumn{3}{c}{\textbf{AVHBench}} \\
\cmidrule(lr){2-5}
\cmidrule(lr){6-8}
 & \textbf{Visual Dom.} & \textbf{Audio Dom.} & \textbf{Language Dom.} & \textbf{Overall Acc.} & 
   \makecell{\textbf{Video-Driven}\\ \textbf{Audio Hall.}} & \makecell{\textbf{Audio-Driven}\\ \textbf{Video Hall.}} & \textbf{Overall Acc.} \\
\midrule
VideoLLaMA2-AV & 71.8 & 80.0 & 68.8 & 73.5 & 75.7 & 79.0 & 77.4 \\
VideoLLaMA2-AV + $\text{VCD}_{\text{Extended}}$ & 71.3 & 83.3 & 74.8 & 76.4 & 66.0 & 74.8 &  70.4 \\
VideoLLaMA2-AV + AVCD & 71.8 & 84.0 & 71.5 & 75.8 & 78.3 & \textbf{80.3} &  79.3 \\
\rowcolor{gray!15} VideoLLaMA2-AV + MAD & \textbf{82.3} & \textbf{84.3} & \textbf{77.5} & \textbf{81.3} & \textbf{79.7} & 79.1 & \textbf{79.4} \\ 
\midrule
Qwen2.5-Omni-7B & 64.5 & 72.3 & 81.3 & 72.7 & 73.0 & 80.7 & 76.9 \\
Qwen2.5-Omni-7B + $\text{VCD}_{\text{Extended}}$ & 62.5 & 71.3 & \textbf{84.5} & 72.8 & 70.3 & 77.1 & 73.7 \\
Qwen2.5-Omni-7B + AVCD & 66.3 & 72.8 & 81.0 & 73.3 & 75.8 & 79.7 & 77.8 \\
\rowcolor{gray!15} Qwen2.5-Omni-7B + MAD & \textbf{76.8} & \textbf{84.3} & 83.3 & \textbf{81.4} & \textbf{78.7} & \textbf{84.4} & \textbf{81.6} \\
\bottomrule
\end{tabular}
}
\caption{Main results on cross-modal hallucination benchmarks. We compare the original decoding (Base), VCD-Extended (with Eq~\eqref{vcd_extended}), AVCD, and our proposed MAD. MAD consistently outperforms all baselines across three audio-visual LLMs and both benchmarks, validating the effectiveness of query-specific modality selection for mitigating cross-modal hallucinations. All results are reported in accuracy (\%).}
\label{tab:main_results}
\end{table*}



%% file: tables/table2_1.tex
\begin{table}[t]
\centering
\caption{Comparison of different modality fusion strategies in the ablation study using VideoLLaMA2-AV as the base model. The proposed adaptive weighting (Weighted) achieves the best overall accuracy across CMM benchmark.}
\renewcommand{\arraystretch}{1.0}
\renewcommand{\tabcolsep}{2.0mm}
\resizebox{1.0\linewidth}{!}{
\begin{tabular}{lcccc}
\toprule
\multirow{2.5}{*}{\makecell{\textbf{Fusion}\\\textbf{Method}}} &
\multicolumn{4}{c}{\textbf{CMM}} \\
\cmidrule(lr){2-5}
 & \textbf{Visual Dom.} & \textbf{Audio Dom.} & \textbf{Language Dom.} & \textbf{Overall Acc.} \\
\midrule
Baseline & 71.8 & 80.0 & 68.8 & 73.5 \\
Uniform & 77.5 & 83.3 & \textbf{77.5} & 79.4 \\
Argmax & 78.5 & 80.5 & 77.0 & 78.7\\
\rowcolor{gray!15} Weighted & \textbf{82.3} & \textbf{84.3} & \textbf{77.5} & \textbf{81.3} \\ 
\bottomrule
\end{tabular}
}
\label{tab:ablation}
\end{table}

%% file: tables/table4.tex
\begin{table}[t]
\centering
\caption{Ablation on modality-specific weights in MAD using VideoLLaMA2-AV as the base model. We disable each weight individually to evaluate its contribution to mitigating cross-modal hallucinations. Using all three weights yields the best overall accuracy.}
\renewcommand{\arraystretch}{1.0}
\resizebox{1.0\linewidth}{!}{
\begin{tabular}{cccccccc} 
\toprule
\multirow{3.5}{*}{\textbf{Decoding}} & \multicolumn{3}{c}{\textbf{Adapted Weights}} & \multicolumn{3}{c}{\textbf{CMM}}& \multirow{3.5}{*}{\textbf{Acc$\uparrow$}} \\
\cmidrule(lr){2-4}
\cmidrule(lr){5-7}
& $w_a$ & $w_v$ & $w_{av}$ & \makecell{\textbf{Visual}\\ \textbf{Dom.}} & \makecell{\textbf{Audio}\\ \textbf{Dom.}} &\makecell{\textbf{Language}\\ \textbf{Dom.}} \\
\midrule
MAD & & \checkmark & \checkmark & 75.8 & 83.5 & 74.8 & 78.0 \\
MAD & \checkmark & & \checkmark & 81.0 & 81.3 & 72.8 & 78.3 \\
MAD & \checkmark & \checkmark & & 80.3 & 81.8 & 74.8 & 78.9 \\
MAD & \checkmark & \checkmark & \checkmark & \textbf{82.3} & \textbf{84.3} & \textbf{77.5} & \textbf{81.3} \\
\bottomrule
\end{tabular}
}
\label{tab:ablation_weight}
\end{table}


%% file: tables/table3.tex
\begin{table}[t]
\centering
\caption{Comparison of MLLMs on general AVQA benchmarks (OmniBench~\cite{li2025omnibenchfutureuniversalomnilanguage}, Worldsense~\cite{hong2025worldsenseevaluatingrealworldomnimodal}, and MUSIC-AVQA~\cite{li2022learning}). All results are reported in accuracy (\%).}
\resizebox{1.0\linewidth}{!}{
\renewcommand{\arraystretch}{0.9}
\renewcommand{\tabcolsep}{2.0mm}
\begin{tabular}{lccc}
\toprule
\textbf{Model} & \textbf{OmniBench} & \textbf{Worldsense} & \textbf{Music-AVQA}
\\
\midrule
VideoLLaMA2-AV & 36.3 & 23.3 & 78.1  \\
VideoLLaMA2-AV + MAD & 36.8 & 25.6 & 79.1  \\
\bottomrule
\end{tabular}
}
\label{tab:general_QA}
\end{table}

%% file: sec/5_conclusion.tex
\section{Conclusion}
In this work, we addressed cross-modal hallucinations in audio-visual large language models by introducing Modality-Adaptive Decoding (MAD), a simple and training-free decoding strategy that aligns contrastive signals with the modality requirements of each question. 
MAD employs a self-assessment mechanism to evaluate modality relevance and assign modality-specific weights, adaptively modulating multi modalities to suppress interference from irrelevant modalities and mitigate hallucination.
Our experiments, demonstrates that MAD effectively suppresses cross-modal hallucinations, suggesting that applying modality-appropriate contrastive decoding contributes to more reliable multimodal reasoning.
Although MAD currently extracts modality weights directly from the underlying MLLM, future work will focus on learning a lightweight, parameter-efficient predictor that estimates modality weights more quickly and accurately. Furthermore, we plan to extend the modality-adaptive framework to richer modality combinations beyond audio
–video, including thermal–RGB or other multi-sensor settings, broadening the impact of modality-aware decoding in real-world multimodal systems.

%% file: sec/X_suppl.tex
\clearpage
\setcounter{page}{1}
\maketitlesupplementary

\section{Impact of $\gamma$}
\label{sec:rationale}

In our modality-adaptive weighting mechanism, the temperature parameter $\gamma$ is a crucial hyperparameter that controls the contrastive strength between modality-specific distributions and the full distribution. To determine the appropriate $\gamma$ value, we conduct a systematic ablation study by varying $\gamma$ from 0.5 to 3.0 with intervals of 0.5. For each $\gamma$ value, we evaluate the performance of our MAD approach on AVHBench~\cite{sung2024avhbench} and CMM~\cite{leng2024curse} benchmarks using VideoLLaMA2-AV~\cite{cheng2024videollama} and Qwen2.5-Omni~\cite{xu2025qwen2} as the base model.

\subsection{Results and Analysis}

\input{figures/suppl_figure1}

Based on this analysis, we use $\gamma = 2.5$ as the default value across all experiments. This value demonstrates consistently robust performance across different models and benchmarks.

\section{Qualitative Analysis of Modality-Adaptive Weights}
\label{sec:modality_weight_analysis}

In the main paper, we present a quantitative analysis demonstrating that the extracted modality weights $w_m$ accurately reflect the modality relevance implied by each task. Here, we provide additional details on the experimental setup and present qualitative examples to further illustrate the behavior of our modality-adaptive weighting mechanism.

\subsection{Dataset Construction}

We randomly sample 100 videos from VideoMME~\cite{fu2025videommefirstevercomprehensiveevaluation} and construct 300 questions categorized into three types based on their modality dependency:

\begin{itemize}
    \item \textbf{Visual-related questions} (100 questions): Questions that require only visual information to answer.
    \item \textbf{Audio-related questions} (100 questions): Questions that require only audio information to answer.
    \item \textbf{Audio-visual-related questions} (100 questions): Questions that require both audio and visual information to answer.
\end{itemize}

\subsection{Question Examples}

Table~\ref{tab:question_examples} presents representative examples of questions from each category along with their corresponding computed modality weights.

\input{tables/table_suppl1}

\subsection{Analysis}

As shown in Table~\ref{tab:question_examples}, our modality-adaptive weighting mechanism successfully assigns higher weights to the relevant modality for each question type. For visual-related questions, $w_v$ is consistently higher, indicating strong reliance on video information. Conversely, audio-related questions exhibit higher $w_a$ values. For audio-visual questions requiring multimodal reasoning, $w_{av}$ receives the highest weight, demonstrating the model's ability to recognize the need for integrated multimodal understanding.

\section{Modality Weight Distribution Analysis Across Tasks}
\label{sec:weight_analysis}

To validate that our modality-adaptive weighting mechanism appropriately adjusts weights according to task characteristics, we analyze the distribution of modality weights across different question types in AVHBench and CMM benchmarks. All experiments are conducted using VideoLLaMA2-AV as the base model.

\subsection{Weight Analysis on AVHBench}

AVHBench is specifically designed to evaluate cross-modal hallucinations by presenting questions where one modality can mislead the model's understanding of another modality. Figure~\ref{fig:avhbench_weights} shows the distribution of modality weights ($w_a$, $w_v$, $w_{av}$) across different task categories.

\input{figures/fig_suppl2_1}

\paragraph{Video-Driven Audio Hallucination (V→A).} In this task category, visual information can mislead audio-related understanding. Despite the potential interference from video, the questions fundamentally require audio comprehension. Our analysis reveals that the model correctly identifies audio as the primary modality, assigning the highest proportion to $w_a$. This demonstrates that MAD successfully prioritizes the relevant modality even in the presence of misleading cross-modal information.

\paragraph{Audio-Driven Video Hallucination (A→V).} Conversely, this task involves audio information potentially interfering with video understanding. The weight distribution shows a predominant $w_v$ proportion, indicating that the model appropriately recognizes video as the essential modality for answering these questions. 

This symmetric behavior across V→A and A→V tasks validates the adaptability of our weighting mechanism.

\subsection{Weight Analysis on CMM}

The CMM benchmark categorizes cross-modal hallucinations into three types based on the dominant modality that causes interference. Figure~\ref{fig:cmm_weights} presents the modality weight distributions for each category.

\input{figures/fig_suppl2_2}

\paragraph{Visual Dominance (Visual Dom.).} This category addresses cases where visual information is over-relied upon, leading to the neglect of audio information. Our analysis shows that the model assigns higher weight to $w_a$, effectively preventing audio hallucinations by emphasizing the underrepresented audio modality.

\paragraph{Audio Dominance (Audio Dom.).} In contrast, this category involves audio information inappropriately influencing video understanding. The weight distribution demonstrates increased $w_v$ proportion, which helps suppress video hallucinations caused by audio interference. This behavior mirrors the Visual Dom. pattern but in the opposite direction, further confirming the mechanism's adaptability.

\paragraph{Language Dominance (Language Dom.).} This category predominantly contains video-related questions that are susceptible to language bias, such as "Did you see the shape of the wheel is circular in the video?" These questions can be answered through linguistic priors without genuine visual understanding. The model compensates for this by assigning substantially higher weight to $w_v$, forcing greater reliance on actual visual evidence and thereby mitigating language-based shortcuts.

\subsection{Discussion}

The consistent patterns observed across both benchmarks demonstrate that our modality-adaptive weighting mechanism successfully captures task-specific modality requirements. The model autonomously adjusts weights to emphasize the most relevant modality while suppressing potentially misleading cross-modal influences, validating the effectiveness of our approach in addressing cross-modal hallucinations.





\section{Robustness Analysis of Modality Query Prompts}

To evaluate the robustness of our modality-adaptive weighting mechanism, we conduct ablation studies with various modality query prompts. While our main paper uses the prompt $X_m =$ \textit{``To answer this question, which modality is needed (audio, video, or both''}, we investigate whether alternative prompt formulations affect the model's ability to generate appropriate modality weights.

\subsection{Alternative Prompt Formulations}

We design the following alternative modality query prompts that maintain similar semantic intent but differ in phrasing:

\begin{itemize}
    \item $X_{m'_1}$: \textit{``Identify which modality is required to answer the question (audio, video, or both)''}
    \item $X_{m'_2}$: \textit{``Given this question, select the necessary modality for reasoning (audio, video, or both)''}
    \item $X_{m'_3}$: \textit{``Which modality does this question require (audio, video, or both)''}
    \item $X_{m'_4}$: \textit{``State the modality relevant for answering this question (audio, video, both)''}
\end{itemize}

\subsection{Experimental Results}
\input{figures/fig_suppl3}
Figure~\ref{fig:prompt_robustness} presents the performance statistics across different modality query prompts on both AVHBench and CMM benchmarks. The results demonstrate remarkable consistency across prompt variations, with standard deviations of only 0.26\% on AVHBench and 0.31\% on CMM. The narrow range between minimum and maximum performance values (0.59\% and 0.83\% respectively, indicated by the vertical bars) further confirms that our approach is highly robust to variations in prompt formulation.

This robustness suggests that the model's modality weight generation relies on semantic understanding of the modality dependency concept rather than specific prompt engineering or surface-level pattern matching. The consistent performance across diverse phrasings validates that MAD captures a fundamental principle of modality-task alignment, making it a principled and reliable approach for addressing cross-modal hallucinations.



\section{Computational Analysis}
\label{sec:computational_analysis}

Our MAD method extracts modality-adaptive weights through self-assessment prompts and applies weighted contrastive decoding during inference. To analyze the computational overhead introduced by our approach, we compare the decoding latency (ms/token)$\downarrow$ with other contrastive decoding-based methods. We conduct all experiments on 8 NVIDIA RTX A6000 GPUs using VideoLLaMA2-AV and Qwen2.5-Omni.
\input{tables/table_suppl2}


\section{Qualitative Results}

We provide quantitative results demonstrating the effectiveness of MAD in Fig~\ref{fig:qual1} and Fig~\ref{fig:qual2}. Here, we provide qualitative examples showing how our method mitigates cross-modal hallucinations.

\newpage

\input{figures/fig_suppl3_1}
\input{figures/fig_suppl3_2}

%

%% file: figures/suppl_figure1.tex
\begin{figure}[H]
\centering
\includegraphics[width=0.9\linewidth]{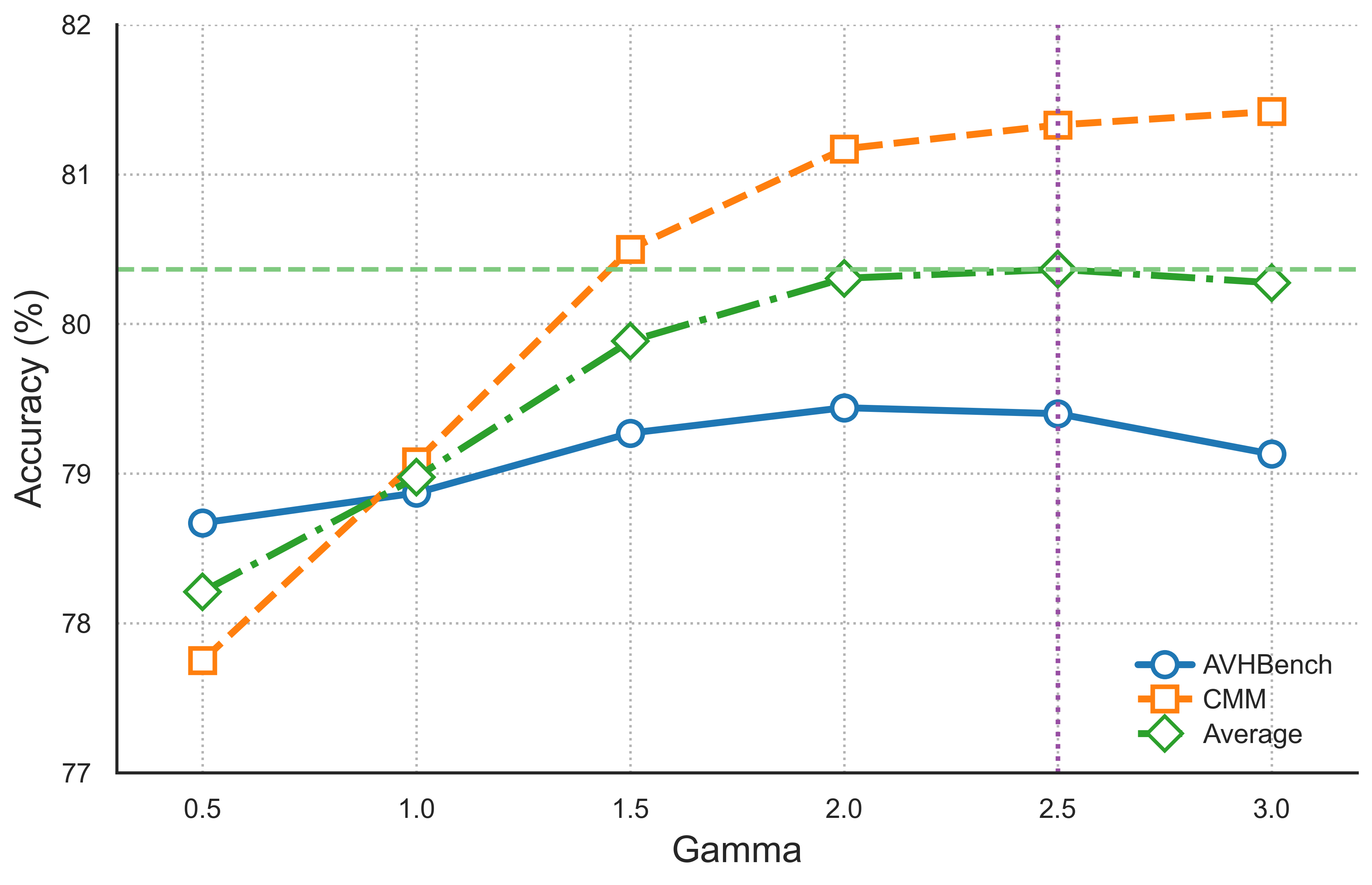}
\caption{Impacts of $\gamma$ in VideoLLaMA2-AV}
\vspace{2mm} 

\includegraphics[width=0.9\linewidth]{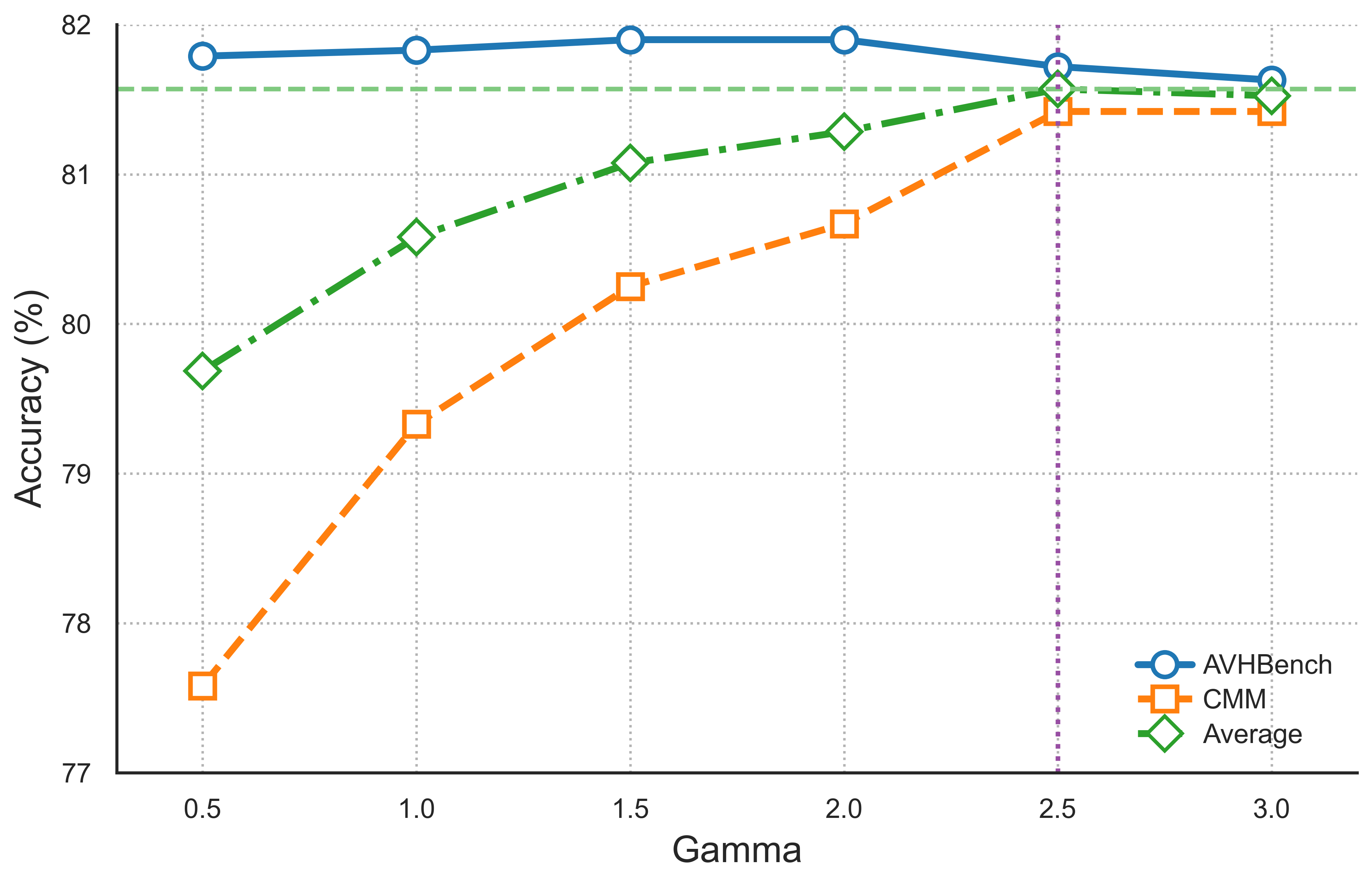}

\vspace{-2mm}
\caption{Impacts of $\gamma$ in Qwen2.5-Omni}
\label{fig:suppl1}
\vspace{-4mm}
\end{figure}

%% file: tables/table_suppl1.tex
\begin{table*}[t]
\centering
\caption{Representative examples of questions and their modality weights.}
\label{tab:question_examples}
\begin{tabular}{llccc}
\toprule
Category & Question & $w_v$ & $w_a$ & $w_{av}$ \\
\midrule
 & Is the singer a black woman? & 0.58 & 0.17 & 0.25 \\
Visual & Does the man with white hair wear glasses? & 0.72 & 0.02 & 0.26 \\
 & Does the character's surroundings get brighter when the character moves? & 0.84 & 0.15 & 0.01 \\
\midrule
 & Can you hear seaguls? & 0.22 & 0.53 & 0.25 \\
Audio & What kind of music is playing in the background? & 0.16 & 0.59 & 0.25 \\
 & Is the background music upbeat and fast-paced? & 0.09 & 0.73 & 0.18 \\
\midrule
 & Is there music when she appears on stage? & 0.27 & 0.16 & 0.56 \\
Audio-Visual & Does the narration match the subtitles? & 0.22 & 0.11 & 0.67 \\
 & Does the character move when there is a clicking sound? & 0.39 & 0.11 & 0.50 \\
\bottomrule
\end{tabular}
\end{table*}

%% file: figures/fig_suppl2_1.tex
\begin{figure}[H]
\centering
\centerline{\includegraphics[width=1.0\linewidth]{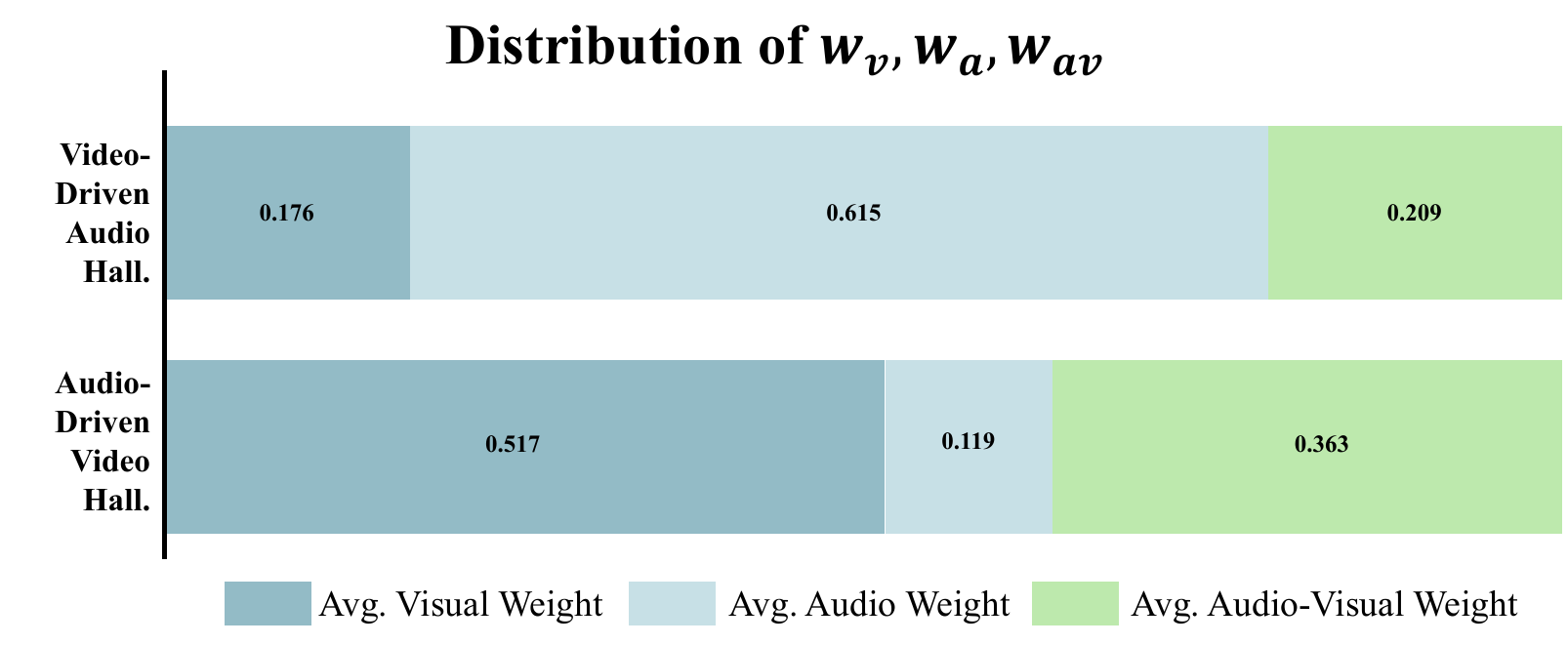}}
\vspace{-2mm}
\caption{Analysis on AVHBench}
\label{fig:avhbench_weights}
\vspace{-4mm}
\end{figure}

%% file: figures/fig_suppl2_2.tex
\begin{figure}[H]
\centering
\centerline{\includegraphics[width=1.0\linewidth]{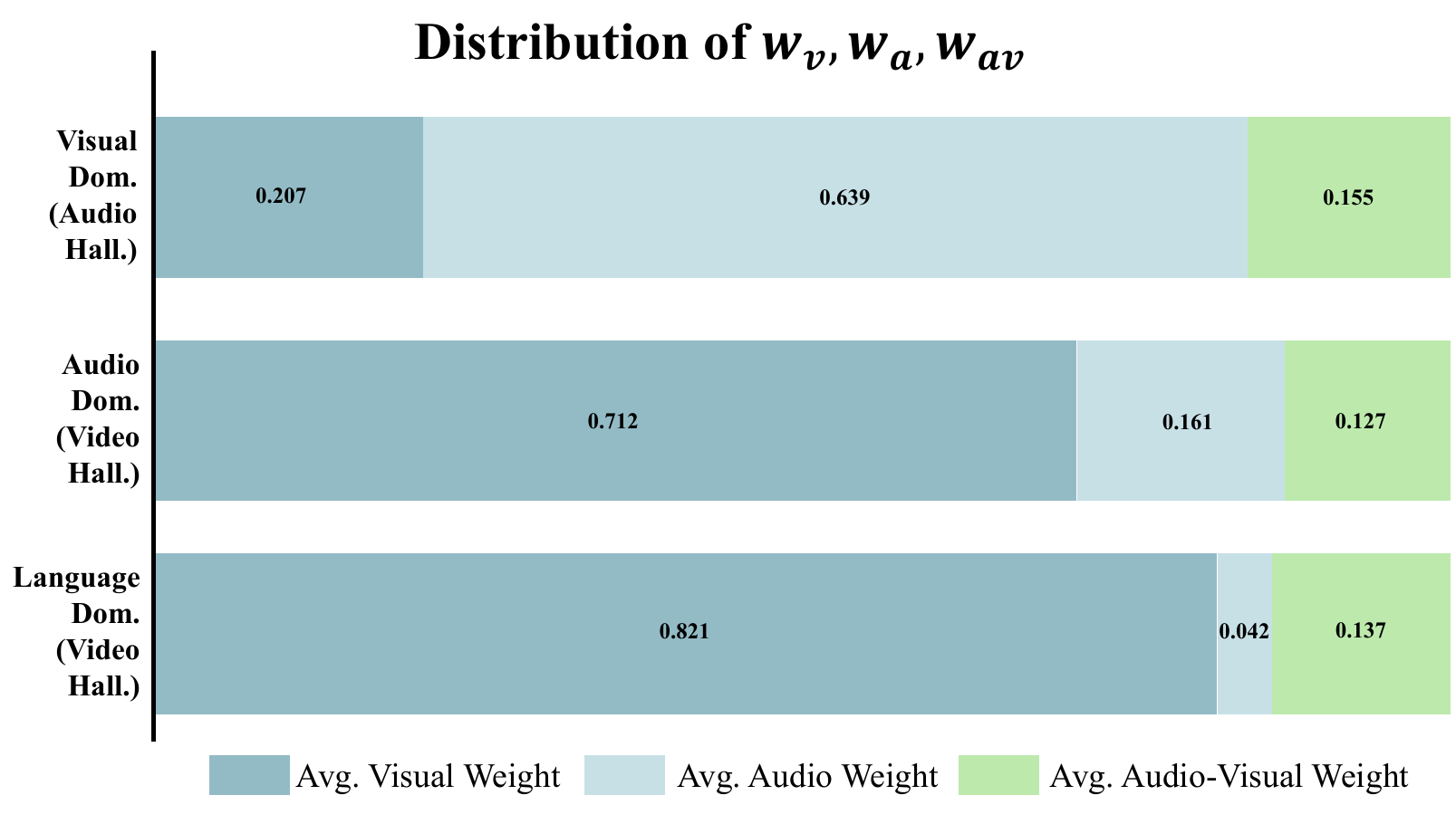}}
\vspace{-2mm}
\caption{Analysis on CMM}
\label{fig:cmm_weights}
\vspace{-4mm}
\end{figure}

%% file: figures/fig_suppl3.tex
\begin{figure}[h]
\centering
\includegraphics[width=1\linewidth]{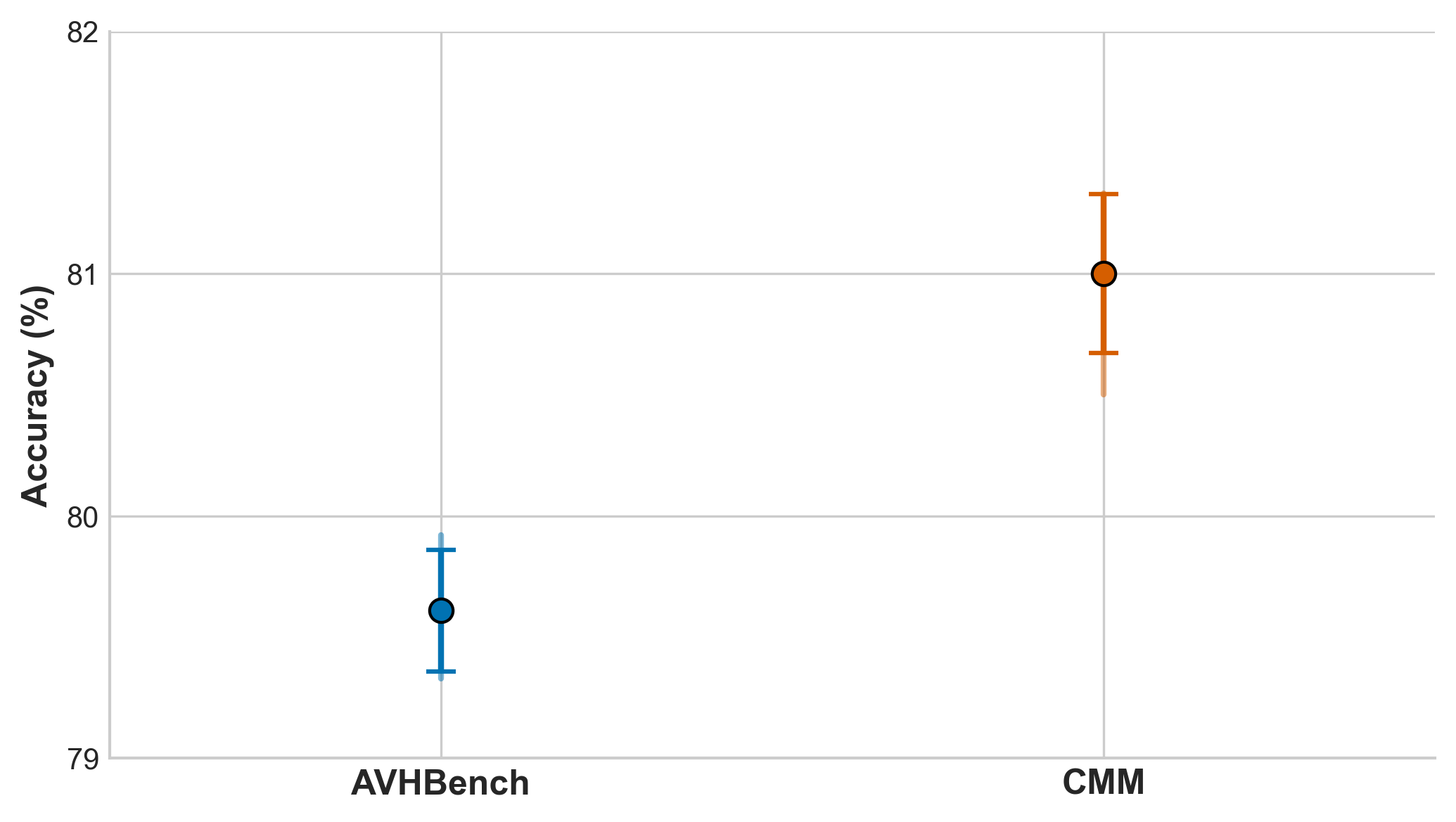}
\caption{Performance consistency across different modality query prompts on AVHBench and CMM benchmarks using VideoLLaMA2-AV. Circles denote mean $\pm$ standard deviation; vertical bars show minimum and maximum accuracy across all five prompt variants.}
\label{fig:prompt_robustness}
\end{figure}

%% file: tables/table_suppl2.tex
\begin{table}[h]
\centering
\caption{Computational efficiency comparison among CD-based methods.}
\label{tab:computational_cost}
\begin{tabular}{lcc}
\toprule
Method & VideoLLaMA2-AV & Qwen2.5-Omni \\
\midrule
VCD$_{\text{extend}}$ & 3564.46 & 4431.78 \\
AVCD & 4811.13 & 9489.81 \\
MAD (Ours) & 3571.64 & 6700.56 \\
\bottomrule
\end{tabular}
\end{table}

%% file: figures/fig_suppl3_1.tex
\begin{figure*}[t]
\centering
\centerline{\includegraphics[width=1.0\linewidth]{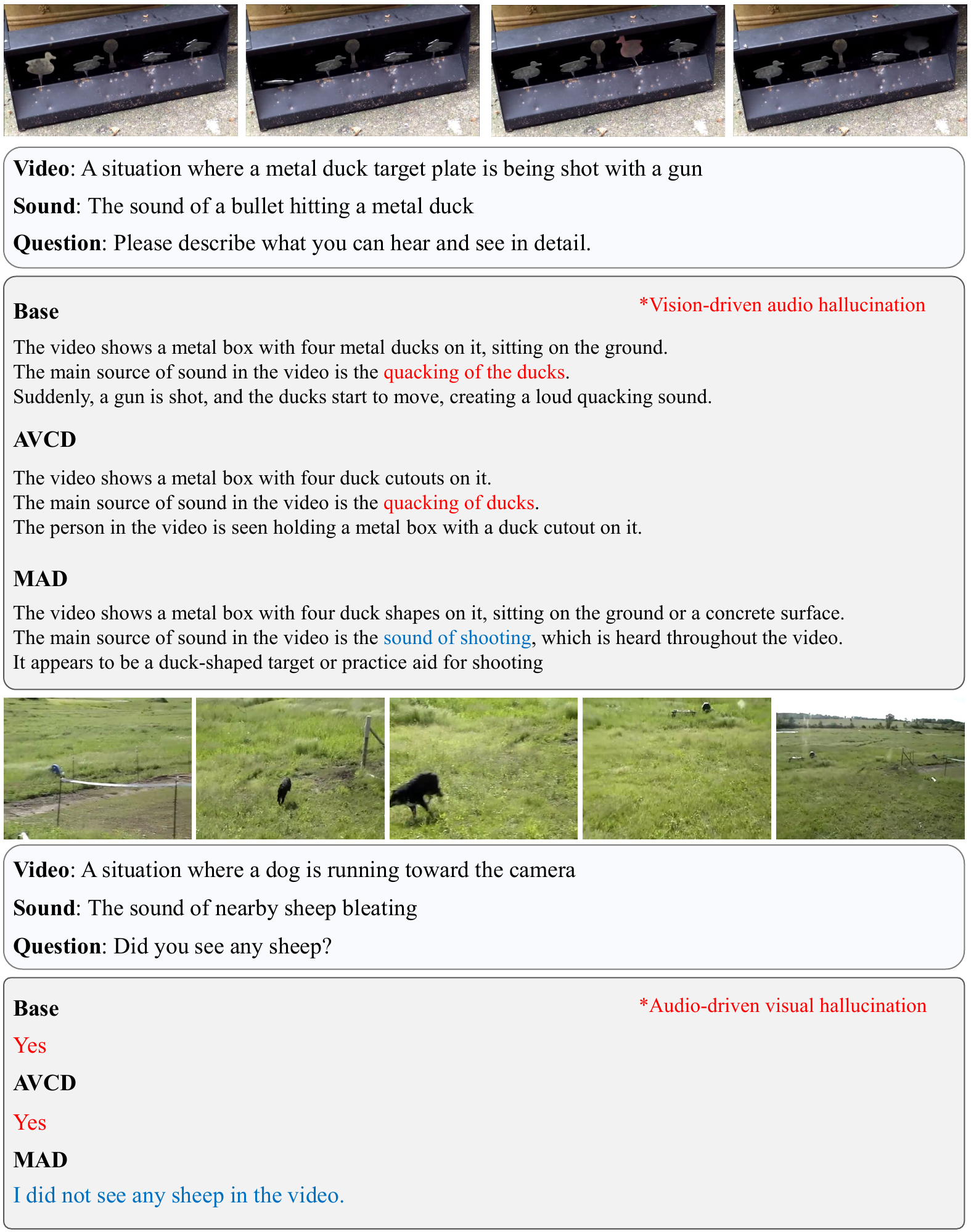}}
\vspace{-2mm}
\caption{Qualitative Results in VideoLLaMA2-AV}
\label{fig:qual1}
\vspace{-4mm}
\end{figure*}

%% file: figures/fig_suppl3_2.tex
\begin{figure*}[t]
\centering
\centerline{\includegraphics[width=0.95\linewidth]{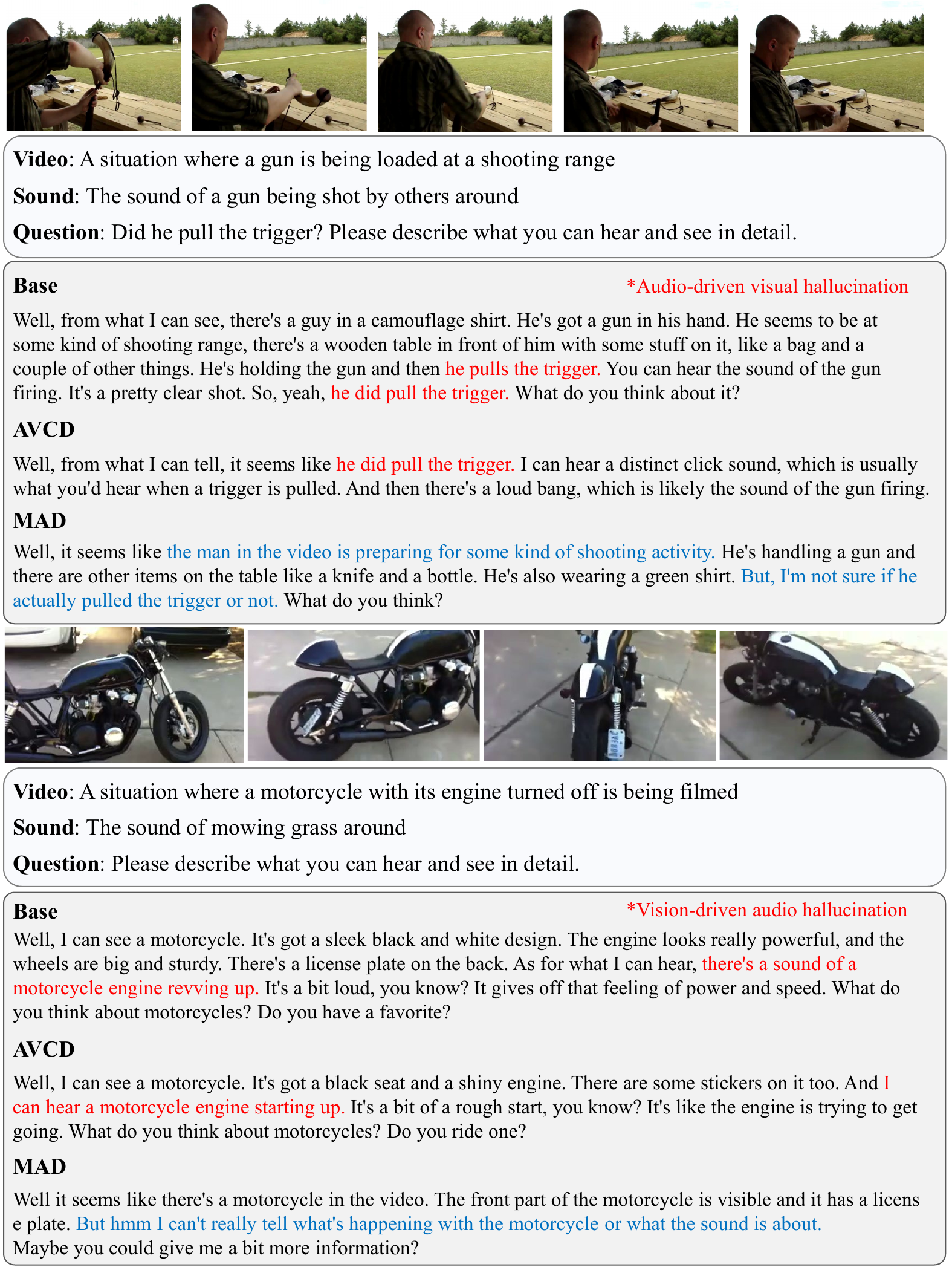}}
\vspace{-2mm}
\caption{Qualitative Results in Qwen2.5-Omni}
\label{fig:qual2}
\vspace{-4mm}
\end{figure*}